\documentclass[10pt,twocolumn,letterpaper]{article}
\pdfoutput=1
\usepackage{cvpr}
\usepackage{times}
\usepackage{epsfig}
\usepackage{graphicx}
\usepackage{amsmath}
\usepackage{amssymb}

\usepackage{xcolor}
\usepackage{booktabs}

\definecolor{ColorGray}{RGB}{128, 128, 128}
\definecolor{ColorDarkGreen}{RGB}{19, 179, 50}
\definecolor{ColorLightBlue}{RGB}{18, 137, 255}

\providecommand{\PaperStatus}{1}

\ifnum \PaperStatus > 0 % Final version

\newcommand{\rev}[1]{}
\def\OurNet{Sketch-R2CNN}
\else % Draft version

\newcommand{\rev}[1]{{\color{ColorGray} [#1]}}
\def\OurNet{{\color{magenta} Sketch-R2CNN}}
\fi % End

\def\Figure{Fig.}

\def\Etal{et al.}
\def\Section{Sec.}

\def\VecSketch{\mathbf{S}}
\def\VecSketchPoint{\mathbf{p}}
\def\VecSketchPointNum{n}
\def\VecSketchPointState{s}
\def\VecSketchPointAttend{a}
\def\RasterSketch{\mathbf{I}}
\def\StrokeWidth{\epsilon}
\def\PixelProjectRatio{\alpha}
\def\LossFunc{L}
\def\BackPropError{\delta^{\RasterSketch}}

\usepackage[pagebackref=true,breaklinks=true,letterpaper=true,colorlinks,bookmarks=false]{hyperref}

\cvprfinalcopy % *** Uncomment this line for the final submission

\def\cvprPaperID{***} % *** Enter the CVPR Paper ID here

\ifcvprfinal\fi

\begin{document}

%%%%%%%%% TITLE
\title{Sketch-R2CNN: An Attentive Network for Vector Sketch Recognition}

\author{Lei Li\\
HKUST\\
\and
Changqing Zou\\
University of Maryland, College Park\\
\and
Youyi Zheng\\
Zhejiang University\\
\and
Qingkun Su\\
Alibaba A.I. Labs\\
\and
Hongbo Fu\\
City University of Hong Kong\\
\and
Chiew-Lan Tai\\
HKUST\\
}

\maketitle
\thispagestyle{empty}

\begin{abstract}
Freehand sketching is a dynamic process where points are sequentially sampled and grouped as strokes for sketch acquisition on electronic devices. To recognize a sketched object, most existing methods discard such important temporal ordering and grouping information from human and simply rasterize sketches into binary images for classification. In this paper, we propose a novel single-branch attentive network architecture \emph{RNN-Rasterization-CNN} (\emph{\OurNet{}} for short) to fully leverage the dynamics in sketches for recognition. \OurNet{} takes as input only a vector sketch with grouped sequences of points, and uses an RNN for stroke attention estimation in the vector space and a CNN for 2D feature extraction in the pixel space respectively. To bridge the gap between these two spaces in neural networks, we propose a neural line rasterization module to convert the vector sketch along with the attention estimated by RNN into a bitmap image, which is subsequently consumed by CNN. The neural line rasterization module is designed in a differentiable way to yield a unified pipeline for end-to-end learning. We perform experiments on existing large-scale sketch recognition benchmarks and show that by exploiting the sketch dynamics with the attention mechanism, our method is more robust and achieves better performance than the state-of-the-art methods. 

\end{abstract}

%%%%%%%%% BODY TEXT
\section{Introduction}
\label{sec:Introduction}

Freehand sketching is an easy and quick means of communication because of its simplicity and expressiveness. 
While a human has the innate ability to interpret drawing semantics, the vast capacity of expressiveness in sketches poses great perception challenges to machines.
For better human-computer interactions,
sketch analysis has been an active research topic in the computer vision and graphics fields, spanning a wide spectrum including 
sketch recognition~\cite{Eitz:2012:HSO,Yu:2017:SketchNet,Zhang:2018:CSC:3229147.3229154}, 
sketch segmentation~\cite{sun2012free,Huang:2014:DSL:2661229.2661280,Li_2018_ECCV,Li:2018:FSS},
sketch-based retrieval~\cite{Eitz:2012:SSR,Wang_2015_CVPR,Sangkloy:2016:Sketchy,Xu_2018_CVPR} 
and modeling~\cite{Olsen:2009:SBM}, etc.
{In this paper, we focus on developing a novel learning-based method for freehand sketch recognition.}

The goal of sketch {classification or} recognition is to identify the object category of an input sketch, which is more challenging than image classification due to the lack of rich texture details, inherent ambiguities, and large shape variations in the input.
Traditional studies~\cite{Eitz:2012:HSO,Schneider:2014:SCC,Li:2015:MKF} commonly cast sketch recognition as an image classification task by converting sketches into binary images and then extracting local image features.
With the quantified feature descriptors, a typical classifier such as Support Vector Machine (SVM) is trained for object category prediction.
Recent years have witnessed the success of deep learning in image classification~\cite{NIPS2012_4824}. 
Similar neural network designs have also been used to address the recognition problem of sketch images~\cite{Yu:2017:SketchNet,Sangkloy:2016:Sketchy}.
Although these deep learning-based methods outperform the traditional ones, the unique properties of sketches, as discussed in the following, are often overlooked, leaving room for further improving the performance of sketch recognition.

\begin{figure*}[t!]
    \centering
    \includegraphics[width=\linewidth]{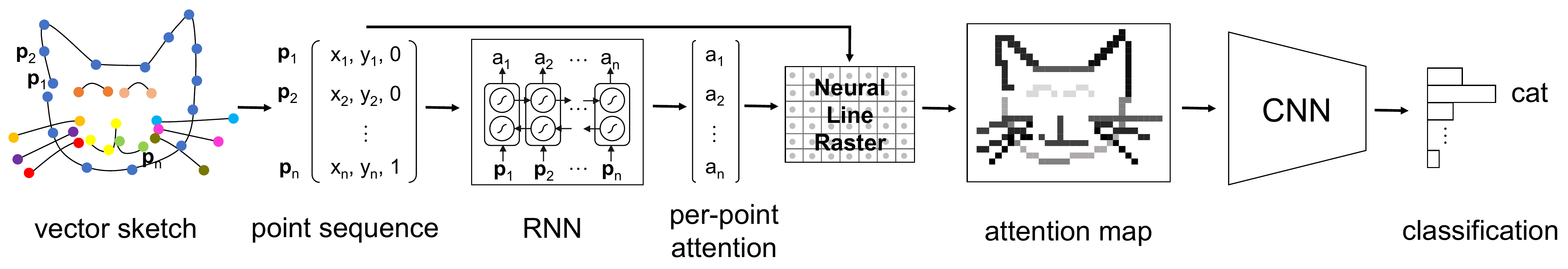}
    \caption{Illustration of our single-branch attentive network architecture for vector sketch recognition. (Neural Line Raster stands for our neural line rasterization (NLR) module.)}
    \label{fig:pipeline}
\end{figure*}

In general, sketch has two widely-used representations for processing, which are raster pixel sketch and vector sketch.
Raster pixel sketches are binary images with pixels covered by strokes {having the value one and the rest of pixels the value zero},
resulting in a large portion of void pixels and thus a sparse representation.
{This representation does not allow the state-of-the-art convolutional neural networks (CNNs) to easily distinguish which strokes are more important or which strokes can be ignored for better recognition~\cite{Schneider:2014:SCC}.}
{Following the definition in~\cite{Xu_2018_CVPR}, a vector sketch in our work refers to a sequence of strokes containing the points in the drawing order}
(\Figure~\ref{fig:pipeline}).
A vector sketch can be easily converted into a bitmap image through rasterization but not vice versa.
{Notably, vector sketches contain rich temporal ordering and grouping (i.e., strokes) information, which has been shown to be useful for learning more descriptive features~\cite{Xu_2018_CVPR}.
However, these information cues are all discarded during the rasterization process for pixel images and thus inaccessible by subsequent recognition algorithms.}

Motivated by the above discussions, to address the incapacity of existing CNN-based methods for stroke importance interpretation,
we propose a novel single-branch attentive network architecture 
\emph{RNN-Rasterization-CNN} (\emph{\OurNet{}} for short), for vector sketch recognition. 
\OurNet{} takes advantages of both vector and raster representations of sketches during the learning process and is able to focus on adaptively learned important strokes, with an attention mechanism, for better recognition (\Figure~\ref{fig:pipeline}). 
It takes only a vector sketch (i.e., grouped sequences of points) as input, and employs a recurrent neural network (RNN) in the first stage for analyzing the {temporal ordering and grouping information} in the input and producing attention estimations for the stroke points.
We then develop a novel \emph{neural line rasterization} (NLR) module, 
capable of converting the vector sketch with the computed attentions into an attention map in a differentiable way.
Subsequently, \OurNet{} uses a CNN to consume the obtained attention map for guided hierarchical understanding and feature extraction on critical strokes to identify the target object category.
Our proposed NLR module is the key to connecting the vector sketch space and the raster sketch space in neural networks and allows gradient information to back propagate from CNN to RNN
for end-to-end learning.
Experiments on existing large-scale sketch recognition benchmarks~\cite{Eitz:2012:HSO,Ha:2017:NS} show that our method, leveraging more human factors in the input, performs better than the state-of-the-art methods,
and our RNN-Rasterization-CNN design consistently improves the performance of CNN-only methods.

In summary, our contributions in this work are: 
(1) the first single-branch attentive network {with an RNN-Rasterization-CNN design} for vector sketch recognition;
(2) a novel differentiable neural line rasterization module that unifies the vector sketch space and raster sketch space in neural networks, allowing end-to-end learning.
We will make our code publicly available.

\section{Related Work}
\label{sec:RelatedWork}

To recognize sketched objects, traditional methods generally take preprocessed raster sketches as input. 
To quantify a sketch image, existing studies have tried to adapt several types of local features originally intended for photos (e.g., bag-of-features~\cite{Eitz:2012:HSO}, Fisher Vectors with SIFT features~\cite{Schneider:2014:SCC},  HOG features~\cite{Li:2015:MKF}) to line drawing images.
With the extracted features, classifiers (e.g., SVMs) are then trained to recognize unseen sketches~\cite{Eitz:2012:HSO,Schneider:2014:SCC}. Different learning schemes, such as multiple kernel learning~\cite{Li:2015:MKF} or active learning~\cite{Yanik:2015:ALSR}, may be employed for performance improvement.
Another line of traditional methods has also attempted to utilize additional cues for recognition, such as prior knowledge for domain-specific sketches~\cite{Alvarado:2004:SMS,LaViola:2004:MSC,Ouyang:2011:CNR,Lu:2005:EAD,sezgin2008sketch,arandjelovic2011sketch} or object context for sketched scenes~\cite{Zhang:2018:CSC:3229147.3229154,Zou_2018_ECCV}.
While progress has been made in sketch recognition, these methods still cannot robustly handle freehand sketches with large {shape or style} variations,
especially those hastily drawn in dozens of seconds~\cite{Ha:2017:NS}, struggling to achieve performance on par with human on existing benchmarks like the TU-Berlin benchmark~\cite{Eitz:2012:HSO}.

% Deep learning based approach
Recently, deep learning has revolutionized many research fields, including sketch recognition, with state-of-the-art performance.
Research efforts~\cite{Sangkloy:2016:Sketchy,Zhang:2016:SSC,Wang:2018:SPN,Yu:2017:SketchNet} have been made to employ deep neural networks, such as AlexNet~\cite{NIPS2012_4824} or GoogLeNet~\cite{Szegedy:2015:Inception}, to learn more discriminative image features in the sketch domain to replace hand-engineered ones.
Yu~\Etal~\cite{Yu:2017:SketchNet} proposed \emph{Sketch-a-Net}, an AlexNet-like architecture specifically adapted for sketch images
by using large kernels in convolutions to accommodate the sparsity of stroke pixels.
Their method achieved superior classification accuracy (77.95\%) on the TU-Berlin benchmark~\cite{Eitz:2012:HSO}, surpassing human performance (73.1\%) for the first time.
Their method still follows the existing learning process of image classification, i.e., using the raster image representation of sketches as CNN inputs, and thus cannot easily learn the awareness of stroke importance in an end-to-end manner for further improvement.
In contrast, our network directly consumes vector sketches as input for  learning stroke importance effectively and adaptively by exploiting the temporal ordering and grouping information therein with RNNs.

Vector representation of sketches has been considered for certain tasks such as sketch generation~\cite{Graves:2013:GSW,Ha:2017:NS,Song_2018_CVPR} or sketch hashing~\cite{Xu_2018_CVPR} with deep learning.
For example, \emph{SketchRNN} 
~\cite{Ha:2017:NS}, which has received much attention recently, is built upon RNNs to process vector sketches.
It is composed of an RNN encoder followed by an RNN decoder, and is able to model the underlying distribution of points in vector sketches for a specific object category.
{To learn to hash sketches for retrieval, Xu \Etal~\cite{Xu_2018_CVPR} has demonstrated that an RNN branch, exploiting temporal ordering in vector sketches, can complement the other CNN branch for extracting more descriptive features.
They fuse two types of features, produced by RNN and CNN respectively, via a late-fusion layer by concatenation.
Our work shares a similar spirit with~\cite{Xu_2018_CVPR}, advocating that the temporal and grouping information in vector sketches also offer additional cues for more accurate sketch recognition.}
In contrast to their two-branch network with simple concatenation, our RNN-Rasterization-CNN design seeks to boost the synergy between the two networks in a single branch during the learning process. 
To this end, inspired by~\cite{Kato_2018_CVPR}, {which proposed an approximate gradient for in-network mesh rendering and rasterization}, we design a novel neural line rasterization module, allowing gradients to back-propagate from CNN (raster sketch space) to RNN (vector sketch space) for end-to-end learning.

{For a sketch, its constituent strokes may contribute differently to its recognition.
With a trained SVM, Schneider \Etal~\cite{Schneider:2014:SCC} qualitatively analyzed how stroke importance affects classification scores by iteratively removing each stroke from the corresponding raster sketch image.
To automatically capture stroke importance during the learning process, researchers have attempted to adapt an attention mechanism in network design~\cite{Song_2017_ICCV}.}
Attention mechanism has been widely used in many visual tasks, such as image classification~\cite{NIPS2014_5542,Xiao_2015_CVPR,Wang_2017_CVPR,Hu_2018_CVPR}, image caption~\cite{XuBKCCSZB15,Lu_2017_CVPR} or Visual Question Answering (VQA)~\cite{Nam_2017_CVPR}.
A simple attention module generally works by computing soft masks over the spatial image grid~\cite{Wang_2017_CVPR,XuBKCCSZB15}, or even feature channels~\cite{Hu_2018_CVPR}, to obtain weighted combination of features.
Song \Etal~\cite{Song_2017_ICCV} has incorporated a spatial attention module for raster sketches in their network for fine-grained sketch-based image retrieval.
Differently, Riaz Muhammad \Etal~\cite{Muhammad_2018_CVPR} tackled the sketch abstraction task with reinforcement learning, which aims to develop a stroke removal policy by considering the stroke influence to recognizability.
{As discussed in existing studies~\cite{Yu:2017:SketchNet,Xu_2018_CVPR,GrahamM17:SSCN,Graham_2018_CVPR}, CNNs may suffer from the sparsity of inputs (e.g., raster sketches), though they excel at building hierarchical representations of 2D inputs.
Instead of struggling to estimate attention from binary images that contain limited information~\cite{Song_2017_ICCV}, we argue that additional cues, such as the temporal ordering and grouping information in vector sketches, are essential to learn reliable attention for strokes.
In our method, we resort to RNNs for computing attention for each point in a vector sketch, and use our NLR module for in-network vector-to-raster conversion.
To our best knowledge, no existing work has tried to derive an attention map from vector sketches with RNNs 
for CNN-based sketch recognition.}

\section{Method}
\label{sec:Method}

Our network architecture, as illustrated in \Figure~\ref{fig:pipeline}, is composed of two cascaded sub-networks:
an RNN for stroke attention estimation in the vector sketch space and a CNN for 2D feature extraction in the raster sketch space (\Section{}~\ref{subsec:NetArch}).
The key enabler for linking the two sub-networks that operate in completely different spaces is a novel neural line rasterization (NLR) module, which converts a vector sketch with the estimated attention to a raster pixel sketch in a differentiable way  (\Section{}~\ref{subsec:Raster}).
More specifically, during the forward inference pass, given a vector sketch as input, the RNN takes in a point at each time step and computes a corresponding attention value for the point.
Our proposed NLR module then rasterizes the vector sketch, together with the estimated per-point attention, into an attention map and computes the corresponding gradients for the backward optimization pass.
A subsequent CNN consumes the attention map
as input for hierarchical understanding and produces category predictions as the final output.

\subsection{Input Representation}
\label{subsec:InputRepresent}
The input to our network is a vector sketch, formed by a sequence of strokes, each stroke being represented by a sequence of points. 
This storing format is widely adopted for sketches in existing crowdsourced datasets~\cite{Ha:2017:NS,Sangkloy:2016:Sketchy,Eitz:2012:HSO}. 

Following~\cite{Graves:2013:GSW}, we denote a vector sketch as an ordered point sequence 
$\VecSketch = \{ \VecSketchPoint_i = (x_i, y_i, \VecSketchPointState_i) \}_{i=1 \cdots \VecSketchPointNum}$, 
where $\VecSketchPointNum$ is the total number of points {in all strokes}.
For each point $\VecSketchPoint_i$, $x_i$ and $y_i$ are the 2D coordinates,
and $\VecSketchPointState_i$ is a binary stroke state.
Specifically, state $\VecSketchPointState_i = 0$ indicates that the current stroke has not ended and that the stroke connects $\VecSketchPoint_i$ to $\VecSketchPoint_{i+1}$; 
$\VecSketchPointState_i = 1$ indicates that $\VecSketchPoint_i$ is the last point of the current stroke and  $\VecSketchPoint_{i+1}$ will be the starting point of another stroke.
Our network takes only the vector sketch $\VecSketch$ as input for end-to-end learning.

\subsection{Network Architecture}
\label{subsec:NetArch}

Our network architecture is formed by two sequentially-arranged sub-networks, which are linked with a differentiable NLR module.
The first sub-network is an RNN, which analyzes the {temporal ordering and grouping} information in the input.
The RNN consumes a vector sketch $\VecSketch$ and estimates per-point attention as output at each iteration step.
Specifically, we use a bidirectional Long Short-Term Memory (LSTM) unit with two layers as the first sub-network.
We set the size of the hidden state to be 512 and adopt dropout with probability = 0.5.
{For the hidden state at step $i$, after the LSTM cell takes in $\VecSketchPoint_i$,  we} pass it through a fully-connected layer followed by a sigmoid function to produce per-point attention, denoted as $\VecSketchPointAttend_i$.
% Attention symbol
That is, for each point $\VecSketchPoint_i$, we obtain a corresponding scalar  $\VecSketchPointAttend_i$, signifying the point importance in the subsequent 2D visual understanding by CNN. 
Similar to~\cite{Ha:2017:NS}, instead of using absolute coordinates, for each $\VecSketchPoint_i$ fed into the RNN, we compute the offsets from its previous point $\VecSketchPoint_{i-1}$ as its coordinates.

Next, we pass the point sequence along with the estimated attention, i.e., $(\VecSketchPoint_i, \VecSketchPointAttend_i)_{i=1 \cdots \VecSketchPointNum}$, through our NLR module, as detailed in~\Section~\ref{subsec:Raster}.
The output of the module is a raster sketch image $\RasterSketch$, which can also be viewed as an attention map with the intensity of each stroke pixel as the corresponding attention.
A deep CNN then takes the image $\RasterSketch$ as input for hierarchical 2D feature extraction.
Sketch-a-Net~\cite{Yu:2017:SketchNet} or ResNet50~\cite{He_2016_CVPR} can be used as the backbone network, which is then connected to a fully-connected layer to produce estimations over all the possible object categories.
We use the cross entropy loss for optimizing the whole network.

{Our network architecture for sketch recognition differs from the one proposed by Xu~\Etal~\cite{Xu_2018_CVPR} for sketch retrieval in several aspects.
First, their network has two branches for feature extraction, one branch with a RNN and the other branch with a CNN.
During learning, their RNN and CNN individually work on two different sketch spaces with little interaction, except at the last concatenation layer for feature fusion.
In contrast, our single-branch design allows more information flow between RNN and CNN owing to our NLR module, that is, the RNN can complement the CNN by producing a more informative input whereas the CNN provides guidance on attention estimation with learned  hierarchical representations during back propagation.
In addition, our network only uses vector sketches as input and performs in-network vector-to-raster conversion, while the two-branch late-fusion network~\cite{Xu_2018_CVPR} requires both vector and raster sketches as input, thus a preprocessing stage for rasterization is needed.
}

\subsection{Neural Line Rasterization with Attention}
\label{subsec:Raster}

\begin{figure}[t!]
    \centering
    \includegraphics[width=0.7\linewidth]{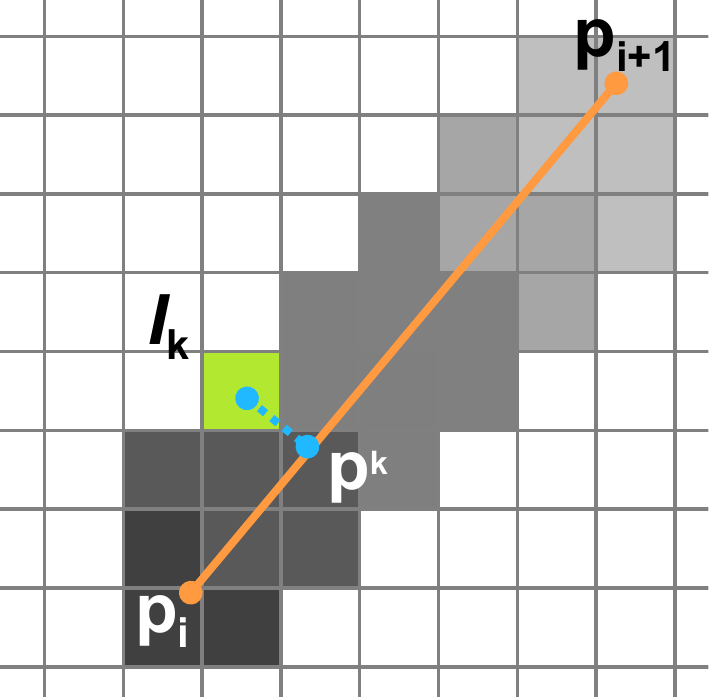}
    \caption{Rasterization of line segment $\VecSketchPoint_i \VecSketchPoint_{i+1}$ and linear interpolation of the attention value for stroke pixel $\RasterSketch_k$.}
    \label{fig:lineraster}
\end{figure}

To convert a point sequence with attention $(\VecSketchPoint_i, \VecSketchPointAttend_i)_{i=1 \cdots \VecSketchPointNum}$ to a pixel image $\RasterSketch$, the basic operation is to draw each valid line segment $\VecSketchPoint_i \VecSketchPoint_{i+1}$ (\Section~\ref{subsec:InputRepresent}) onto the canvas image. 
{As illustrated in \Figure~\ref{fig:lineraster}, to determine whether or not a pixel $\RasterSketch_k$ is on the target line segment, }
we simply compute the distance from its center to the line segment $\VecSketchPoint_i \VecSketchPoint_{i+1}$ 
and check whether it is smaller than a predefined threshold $\StrokeWidth$
(we set $\StrokeWidth = 1$ in our experiments).
If $\RasterSketch_k$ is a stroke pixel, we compute its attention by linear interpolation~\cite{Kato_2018_CVPR}; otherwise its attention is set to zero.
More specifically, let $\VecSketchPoint^k$ be the projection point of $\RasterSketch_k$'s center onto $\VecSketchPoint_i \VecSketchPoint_{i+1}$.
The intensity or attention of $\RasterSketch_k$ is then defined as 
\begin{equation}
\begin{split}
\RasterSketch_k = (1 - \PixelProjectRatio_k) \cdot \VecSketchPointAttend_i + \PixelProjectRatio_k \cdot \VecSketchPointAttend_{i+1} ,
\end{split}
\end{equation}
where $\PixelProjectRatio_k = \| \VecSketchPoint^k - \VecSketchPoint_i \|_2 / \| \VecSketchPoint_{i+1}  - \VecSketchPoint_i \|_2$, and {$\VecSketchPoint^k$, $\VecSketchPoint_i$ and $\VecSketchPoint_{i+1}$ are in absolute coordinates.}
{This rasterization process for line segments can be efficiently done in parallel on GPU with a CUDA kernel.
Note that in the implementation we need to record the relevant information, such as line segment index and $\PixelProjectRatio_k$ at each pixel $\RasterSketch_k$, for subsequent gradient computation.}

% Figure

Through the above process, a vector sketch can be easily converted into a raster image in the forward inference pass.
In order to propagate gradients {w.r.t the loss function} from CNN to RNN in the backward optimization pass, we need to derive gradients for the above rasterization process. 
Thanks to the simplicity of the used linear interpolation, the gradients can be computed as follows:
\begin{equation}
\begin{split}
\dfrac{\partial \RasterSketch_k}{\partial \VecSketchPointAttend_i} = 1 - \PixelProjectRatio_k , \ \
\dfrac{\partial \RasterSketch_k}{\partial \VecSketchPointAttend_{i+1}} = \PixelProjectRatio_k .
\end{split}
\end{equation}
Let $\LossFunc$ be the loss function and $\BackPropError_{k}$ be the gradient back-propagated into $\RasterSketch_k$ w.r.t $\LossFunc$ through the CNN.
By the chain rule, we have
\begin{equation}
\begin{split}
\dfrac{\partial \LossFunc}{\partial \VecSketchPointAttend_i} = \sum_{k} \BackPropError_{k} \cdot (1 - \PixelProjectRatio_k) , \ 
\dfrac{\partial \LossFunc}{\partial \VecSketchPointAttend_{i+1}} = \sum_{k} \BackPropError_{k} \cdot \PixelProjectRatio_k ,
\end{split}
\end{equation}
{where $k$ iterates over all the stroke pixels covered by the line segment $\VecSketchPoint_i \VecSketchPoint_{i+1}$.
If $\VecSketchPoint_i$ is adjacent to another line segment $\VecSketchPoint_{i-1} \VecSketchPoint_{i}$, we accumulate the gradients.

Our NLR module is simple and easy to implement, but it is crucial to bridge the gap between the vector sketch space and the raster sketch space in neural networks for end-to-end learning.
Unlike existing methods~\cite{Wang_2017_CVPR,Song_2017_ICCV} that derive attention from feature maps produced by CNNs, with our NLR module, we can take advantage of additional cues (i.e., temporal ordering and grouping information) in vector sketches for better attention map estimation, as shown in experiments (\Section{}~\ref{subsec:resultdiscuss}).
These additional cues, however, are not accessible for the methods with raster inputs.

\section{Experiments}
\label{sec:Experiments}

%=================================================================================
\subsection{Datasets and Settings}
We have performed various experiments on two existing large-scale sketch recognition benchmarks, i.e., the TU-Berlin benchmark~\cite{Eitz:2012:HSO} 
and the QuickDraw benchmark~\cite{Ha:2017:NS}, to validate the performance of our \OurNet{}.
These two benchmarks differ in several aspects, such as sketching style, acquisition procedure, and sketch quantity per category.
Notably, sketches in the TU-Berlin benchmark tend to be more realistic while the ones in QuickDraw are more iconic and abstract (\Figure~\ref{fig:cnnfail-oursucceed}).
The TU-Berlin benchmark~\cite{Eitz:2012:HSO} contains 250 object categories with 80 sketches per category.
Each sketch was created within 30 minutes by a participant from Amazon Mechanical Turk (AMT).
The QuickDraw benchmark~\cite{Ha:2017:NS} contains 345 object categories with 75K sketches per category.
During acquisition, the participants were given only 20 seconds to sketch an object.

% Sketch processing: RDP for simplification, max number of points.
Similar to~\cite{Ha:2017:NS}, to simplify sketches in the TU-Berlin benchmark, we applied the Ramer-Douglas-Peucker (RDP) algorithm, resulting a maximum point sequence length of 448 for RNN.
{Following~\cite{Yu:2017:SketchNet}, we used three-fold cross validation on this benchmark (i.e., two folds for training, one fold for testing).}
Sketches in the QuickDraw benchmark have already been preprocessed with the RDP simplification algorithm and the maximum number of points in a sketch is 321.
In each QuickDraw category, the 75K sketches have already been divided into training, validation and testing sets with sizes of 70K, 2.5K and 2.5K, respectively.

{We implemented our \OurNet{} and NLR module with PyTorch.
We adopted Adam~\cite{KingmaB:2014:Adam} for stochastic gradient descent update with a mini-batch size of 48.
We used a learning rate of 0.0001 for training on QuickDraw and 0.00005 for training or fine-tuning on TU-Berlin (see \Section~\ref{subsec:resultdiscuss} for the pre-training and training procedures).
Due to the limited training data in the TU-Berlin benchmark, we followed~\cite{Yu:2017:SketchNet} to perform data augmentation, including horizontal reflection, stroke removal and sketch deformation.}

%=================================================================================
\subsection{Results and Discussions}
\label{subsec:resultdiscuss}
\textbf{Results on TU-Berlin Benchmark.}
{We have compared our method with a variety of existing methods  on the TU-Berlin benchmark. Table~\ref{tab:tuberlin} includes the results of some methods reported in~\cite{Yu:2017:SketchNet}.}
These methods can be generally categorized into two groups. The first group follows the conventional pipeline using hand-crafted features + classifier, including the HOG-SVM method~\cite{Eitz:2012:HSO}, structured ensemble matching~\cite{Li:2013:SRE}, multi-kernel SVM~\cite{Li:2015:MKF}, and
the Fisher Vector based method~\cite{Schneider:2014:SCC}. 
The second group uses deep learning, including the state-of-the-art network Sketch-a-Net (the earlier version Sketch-a-Net v1 \cite{YuYSXH15} and the later improved version Sketch-a-Net v2~\cite{Yu:2017:SketchNet}) and those networks that have been evaluated in~\cite{Yu:2017:SketchNet}: LeNet~\cite{LeCun2012}, AlexNet-SVM~\cite{NIPS2012_4824} and AlexNet-Sketch~\cite{NIPS2012_4824}. 

% ------ Table ------> 
%https://www.tablesgenerator.com/

\begin{table}[ht]
\begin{center}
\begin{tabular}{l | c}
	\toprule
	Model & Accuracy \\
	\midrule
	Humans~\cite{Eitz:2012:HSO} & 73.1\% \\
	\midrule
	HOG-SVM~\cite{Eitz:2012:HSO} &  56.0\% \\
	Ensemble~\cite{Li:2013:SRE}  &  61.5\% \\
	MKL-SVM~\cite{Li:2015:MKF} &  65.8\% \\
	Fisher-Vectors~\cite{Schneider:2014:SCC}      &  68.9\% \\
	\midrule
	LeNet~\cite{LeCun2012} & 55.2\% \\
	AlexNet-SVM~\cite{NIPS2012_4824} & 67.1\%  \\
	AlexNet-Sketch~\cite{NIPS2012_4824} & 68.6\% \\
	Sketch-a-Net v1~\cite{YuYSXH15} & 74.9\% \\
	Sketch-a-Net v2~\cite{Yu:2017:SketchNet} & 77.95\% \\
	Sketch-a-Net v2 (ours)~\cite{Yu:2017:SketchNet} & 77.54\% \\
	ResNet50~\cite{He_2016_CVPR}  & 82.08\% \\
	\textbf{\OurNet{} (Sketch-a-Net v2)}        &  \textbf{78.49\%} \\
	\textbf{\OurNet{} (ResNet50)}       &  \textbf{83.25\%} \\
	\bottomrule
\end{tabular}
\end{center}
\caption{Evaluations on the TU-Berlin benchmark. Our method with ResNet50 working as the CNN backbone achieves the highest recognition accuracy. Sketch-a-Net v2 (our) is our PyTorch-based implementation.}
\label{tab:tuberlin}
\end{table}
% <------ Table ------ 

We reimplemented Sketch-a-Net v2 with PyTorch since the original model~\cite{Yu:2017:SketchNet}, implemented with Caffe, is not compatible with our framework (i.e., the NLR module). 
We pre-trained the Sketch-a-Net v2 on QuickDraw~\cite{Ha:2017:NS} instead of preprocessed edge maps from photos~\cite{Yu:2017:SketchNet} for ease of preparation and reproduction.
Our best reproduced recognition accuracy of Sketch-a-Net v2 on the TU-Berlin benchmark is $77.54\%$, close to the accuracy of $77.95\%$ reported with the original Caffe-based implementation~\cite{Yu:2017:SketchNet}. 
{In addition to Sketch-a-Net v2, we also evaluated ResNet50~\cite{He_2016_CVPR}, a more advanced CNN architecture that has been widely used for various visual tasks such as image classification~\cite{He_2016_CVPR} or object detection~\cite{Lin_2017_CVPR}.}
Specifically, before training on raster sketches of the TU-Berlin benchmark, we sequentially pre-trained the ResNet50 on ImageNet~\cite{Russakovsky2015} and QuickQraw. 
The ResNet50 achieves a recognition accuracy of $82.08\%$,  significantly outperforming the state-of-art approach Sketch-a-Net v2. 

Since both Sketch-a-Net v2 and ResNet50 are CNN variants, they can be incorporated into our network architecture (\Figure{}~\ref{fig:pipeline}) as the CNN backbone. 
By inserting one of these CNN alternatives into the proposed architecture, we can study how helpful the attention learned by RNN can be for vector sketch recognition. 
The comparison results are summarized in Table~\ref{tab:tuberlin}. 
Our method incorporated with Sketch-a-Net v2, named
\OurNet{} (Sketch-a-Net-v2) in Table~\ref{tab:tuberlin}, achieves a recognition accuracy of $78.49\%$, improving Sketch-a-Net v2 (ours) by about $1\%$.
Another variant of our method with ResNet50, named \OurNet{} (ResNet50) in Table~\ref{tab:tuberlin}, achieves an accuracy of $83.25\%$, improving the ResNet50-only model by about $1.2\%$, and surpasses all the existing approaches and human performance.

%=================================================================================
\textbf{{Alternatives Study on TU-Berlin Benchmark.}} 
{To validate our proposed architecture, we have studied several network design alternatives on the TU-Berlin benchmark (Table~\ref{tab:ablationTUBerlin}).
First, as mentioned in \Section{}~\ref{sec:RelatedWork}, attention modules have been used in existing CNN architectures for image classification~\cite{Wang_2017_CVPR} and sketch retrieval~\cite{Song_2017_ICCV}.
To compare against our RNN-based attention module, we modified ResNet50 and inserted the spatial attention module proposed by Song \Etal~\cite{Song_2017_ICCV} 
after the $C_2$ residual block~\cite{He_2016_CVPR,Lin_2017_CVPR}.
This modified version of ResNet50 still takes binary sketch images as input and tries to compute attention maps from feature maps of previous convolutional layers.
This model, named Attentive-ResNet50 in Table~\ref{tab:ablationTUBerlin}, achieves a recognition accuracy of $82.42\%$, slightly higher than $82.08\%$ by the ResNet50-only model, while lower than $83.25\%$ attained by our method, showing the comparatively higher effectiveness of additional cues in vector sketches used by our method for attention estimation.
Attention maps produced by our RNN-based attention module and Attentive-ResNet50 are visualized in \Figure{}~\ref{fig:our-resnet50-attention-visualize}.
Note that our method only predicts attention for stroke pixels and sets non-stroke pixels to have an attention value of zero, while Attentive-ResNet50 computes attention for every pixel of the attention map.

% ------ Figure ------> 
\begin{figure}[t!]
    \centering
    \includegraphics[width=\linewidth]{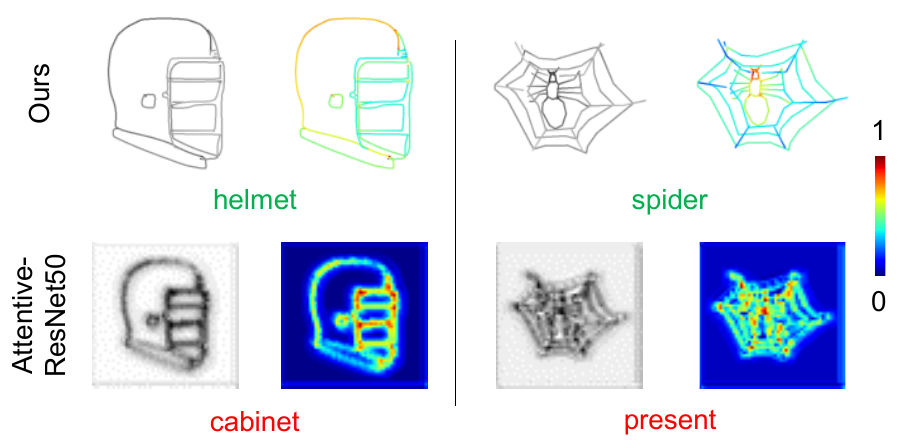}
    \caption{Visualization of attention maps, {in grayscale and color coded}, produced by our \OurNet{} (ResNet50) and Attentive-ResNet50. Recognition failures are in {\color{red} red} and successes are in  {\color{green} green}. Attention maps of Attentive-ResNet50 are estimated from feature maps of the last layer of the $C_2$ residual block, which are of size $56 \times 56$, while attention maps by our method are of size $224 \times 224$. (Best viewed in the electronic version.)}
    \label{fig:our-resnet50-attention-visualize}
\end{figure}
% <------ Figure ------ 

{To study the influence of temporal ordering information provided by human on RNN's attention estimation, we trained \OurNet{} (ResNet50) 
with randomized stroke orders. 
That is, instead of keeping the human drawing order in vector sketch, the stroke sequence is randomly disrupted.}
This scheme, named Random-Stroke-Order, achieves a slightly lower recognition accuracy of $82.78\%$ than \OurNet{} (ResNet50) on the TU-Berlin benchmark, still superior to the ResNet50-only model. 
This indicates that the temporal information (i.e., stroke order) provided by human can help RNN to learn more descriptive sequential features, confirming a similar conclusion made from sketch retrieval experiments in~\cite{Xu_2018_CVPR}.

% ------ Table ------>
%https://www.tablesgenerator.com/

\begin{table}[ht]
\begin{center}
\begin{tabular}{l | c}
	\toprule
	Model & Accuracy \\
	\midrule
    Attentive-ResNet50~\cite{Song_2017_ICCV} & 82.42\% \\
    Random-Stroke-Order     &  82.78\% \\
    Attention-using-Sketching-Order &  81.74\% \\
    Two-Branch-Late-Fusion~\cite{Xu_2018_CVPR} & 81.43\% \\
    Two-Branch-Early-Fusion  &  81.84\% \\
    \textbf{\OurNet{} (ResNet50)}       &  \textbf{83.25\%} \\
	\bottomrule
\end{tabular}
\end{center}
\caption{Alternative design choice studies on the TU-Berlin benchmark.}
\label{tab:ablationTUBerlin}
\end{table}

% <------ Table ------

In addition to our RNN-based encoding method for vector sketches, we also explored a straightforward approach to allow CNNs to gain access to the sketching order information for feature extraction.
Specifically, in a preprocessing step, for a sketch in the point sequence representation, we encode its ordering information into an image through rasterization by assigning an intensity value of one to the first point and zero to the last point and linearly interpolating the intensities of the points in-between.
{\Figure~\ref{fig:attention-visualize} shows some examples of the resulting images.}
This encoding scheme is based on a hypothesis that users tend to draw  more ``important" strokes first, and the resulting raster sketches can be considered as temporal-encoding attention maps.
We trained a ResNet50 with such hand-crafted attention maps as input, but found that this encoding scheme, with a recognition accuracy of 81.74\% (Attention-using-Sketching-Order in Table~\ref{tab:ablationTUBerlin}), is not effective and even slightly worse than the baseline with binary image inputs (ResNet50 in Table~\ref{tab:tuberlin}).
This indicates that, for CNN-based recognition networks, stroke importance may not always be properly aligned with stroke order under such a straightforward encoding scheme, due to different drawing styles used by different users, and this encoding scheme may even pose challenges to CNNs for learning effective patterns.
Thus, instead of ``hard-coding'' temporal information into images, a more adaptive and robust encoder (e.g., RNN) is needed to accommodate sequential variations in vector sketches.

{Next, we discuss arrangements of RNN and CNN in the network architecture design.}
As mentioned before, Xu et al. \cite{Xu_2018_CVPR} use a two-branch late-fusion architecture, which fuses the features extracted from a CNN branch and a parallel RNN branch, for sketch retrieval.
In contrast, our design combines an RNN encoder and a CNN feature extractor sequentially in a single branch for sketch classification. 
We therefore set up another experiment to investigate {which of the above two types of architecture} is a better scheme to incorporate the addition temporal ordering and grouping information existing in vector sketches.
{Following~\cite{Xu_2018_CVPR}, we built a similar model, named Two-Branch-Late-Fusion in Table~\ref{tab:ablationTUBerlin}, which uses the same RNN cell and CNN backbone as \OurNet{} (ResNet50) for fairness and consistency.}
The training procedure is the same as \OurNet{} (ResNet50), with the softmax cross entropy loss~\cite{Xu_2018_CVPR}.
The Two-Branch-Late-Fusion achieves a recognition accuracy of $81.43\%$ on the TU-Berlin benchmark, which is about $2\%$ lower than \OurNet{} (ResNet50). 
This result reveals that our proposed single-branch architecture can make the CNN, which works as an abstract visual concept extractor, and the RNN, which models human sketching orders, complement each other better than the two-branch architecture. 
Surprisingly, another observation is that the recognition accuracy of Two-Branch-Late-Fusion, adapted to the sketch classification task from the original sketch retrieval task,  is even slightly inferior to that of the single CNN branch (ResNet50 in Table~\ref{tab:tuberlin}).
{This is also observed from results} on the QuickDraw benchmark, as presented in the following section. 
Due to the lack of implementation details of~\cite{Xu_2018_CVPR}, we postulate that the differences of training strategies (\cite{Xu_2018_CVPR}: multi-stage training for CNN and RNN; Ours: joint training of CNN and RNN), CNN backbones (\cite{Xu_2018_CVPR}: AlexNet; Ours: ResNet50) and datasets (\cite{Xu_2018_CVPR}: pruned QuickDraw dataset; Ours: original TU-Berlin and QuickDraw datasets) may affect the learning of the late-fusion layer and cause the performance degradation.

% ------ Figure ------> 
\begin{figure*}[h]
    \centering
    \includegraphics[width=\linewidth]{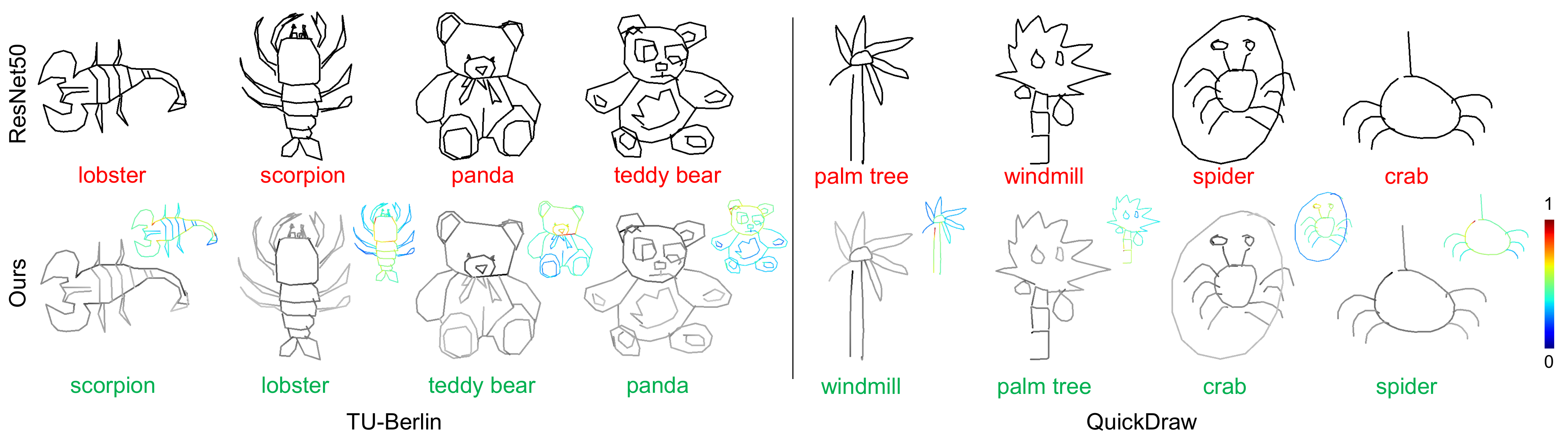}
    \caption{Recognition comparisons between the CNN-only method (ResNet50) and our \OurNet{} (ResNet50 as the CNN backbone). Failures are in {\color{red} red} and successes are in  {\color{green} green}. Attention maps produced by our RNN are shown in the second row and are color coded. {Note that our RNN only predicts attention for stroke pixels; non-stroke pixels are set to have an attention value of zero and are not color-coded.}}
    \label{fig:cnnfail-oursucceed}
\end{figure*}

% <------ Figure ------

% ------ Figure ------> 
\begin{figure*}[h]
    \centering
    \includegraphics[width=\linewidth]{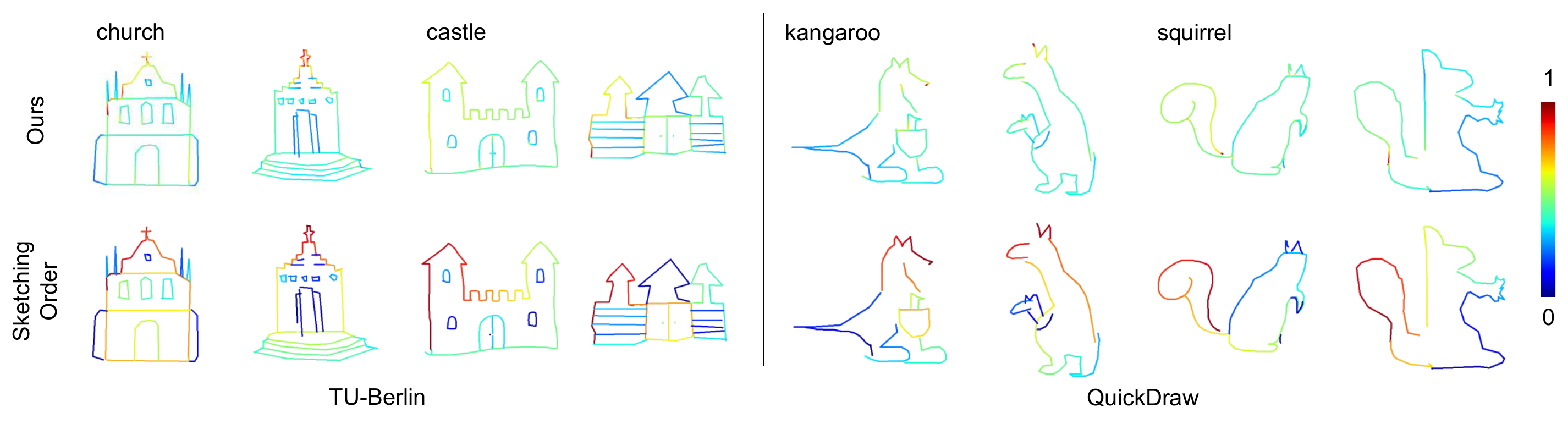}
    \caption{The first row shows color-coded attention maps produced by our \OurNet{} (ResNet50) for specific object categories. Correspondingly in the second row, we directly encode the sketching order as attention maps, higher attention values for strokes drawn earlier. Note that non-stroke pixels are set to have an attention value of zero and are not color-coded.}
    \label{fig:attention-visualize}
\end{figure*}

% <------ Figure ------

Complement to the above experiments on attention estimation with RNN as well as arrangements of RNN and CNN, we stretched the design choice exploration to studying an alternative way of injecting the learned attention from RNN into CNN. 
In our proposed architecture, the CNN directly takes the attention maps produced by the RNN as input. 
An alternative architecture is to weigh feature maps of a certain intermediate layer in CNN (which still takes binary sketch images as input) with the attention map by RNN that leverages vector sketches as input.
In our implementation, we inject the attention map produced by RNN, which is of size $56 \times 56$ with stroke width threshold $\StrokeWidth = 0.5$, into the output of the $C_2$ residual block~\cite{He_2016_CVPR,Lin_2017_CVPR} of ResNet50.
Following the same training procedures as those in Table~\ref{tab:ablationTUBerlin}, 
this alternative architecture, named Two-Branch-Early-Fusion, achieves a recognition accuracy of $81.84\%$ on the TU-Berlin benchmark, performing slightly better than Two-Branch-Late-Fusion. 
However the recognition accuracy of Two-Branch-Early-Fusion is still slightly inferior to that of the ResNet50-only model.
{This may be due to non-stroke pixels in the attention map from RNN having an attention value of zero, which, during the injection, would make convolution features at those corresponding locations vanish, reducing the feature information learned by previous convolutional layers from the input.}

% ------ Table ------> 
%https://www.tablesgenerator.com/

\begin{table}[ht]
\begin{center}
\begin{tabular}{l | c}
	\toprule
	Model & Accuracy \\
	\midrule
	%ours  &  74.84\% \\
	Sketch-a-Net v2~\cite{Yu:2017:SketchNet} & 74.84\% \\
	ResNet50~\cite{He_2016_CVPR}  & 82.48 \% \\
	Two-Branch-Late-Fusion~\cite{Xu_2018_CVPR} & 82.11\% \\
	\textbf{\OurNet{} (Sketch-a-Net v2)}         &  \textbf{77.29\%} \\
	\textbf{\OurNet{} (ResNet50)}        &  \textbf{84.41\%} \\
	\bottomrule
\end{tabular}
\end{center}
\caption{Evaluations on the QuickDraw benchmark.}
\label{tab:quickdraw}
\end{table}

% <------ Table ------

%=================================================================================
\textbf{Results on QuickDraw Benchmark.}
We further compared the proposed \OurNet{} with 
Sketch-a-Net v2~\cite{Yu:2017:SketchNet}, ResNet50-only model, 
and Two-Branch-Late-Fusion~\cite{Xu_2018_CVPR} on the QuickDraw benchmark.
Note the ResNet50 is pre-trained on ImageNet~\cite{Russakovsky2015} and served as the CNN backbone in \OurNet{} and Two-Branch-Late-Fusion.
Quantitative results are summarized in Table~\ref{tab:quickdraw}, 
and the performance of each competing method on the QuickDraw benchmark agrees well with those on the TU-Berlin benchmark. 
Compared to the competitors, \OurNet{} (ResNet50) achieves the highest recognition accuracy on the QuickDraw benchmark, echoing its performance on the TU-Berlin benchmark. It is a similar case for the ResNet50-only model, which still achieves better recognition performance than both Sketch-a-Net v2 and Two-Branch-Late-Fusion. 
\OurNet{} (ResNet50) and \OurNet{} (Sketch-a-Net v2) improve ResNet50 and Sketch-a-Net v2 respectively by about $2\%$. 
Although the sketch quality of QuickDraw may not be as good as that of TU-Berlin, 
thanks to the voluminous data of QuickDraw (24.15M sketches for training, 862.5K sketches for validation or testing), 
we still have seen consistent performance improvement of \OurNet{} over CNN-only models, showing the generality of our proposed architecture.

%=================================================================================
\textbf{Qualitative Results.}
\Figure~\ref{fig:cnnfail-oursucceed} shows some qualitative recognition comparisons between the CNN-only method (ResNet50) and our \OurNet{} (ResNet50).
Through visualization, it is observed that the attention maps produced by the RNN in \OurNet{} can help the CNN to focus on more effective stroke parts of the inputs and ignore the interference of irrelevant strokes (e.g., the circle around the crab in \Figure~\ref{fig:cnnfail-oursucceed}) to make better classifications.
In contrast, the CNN-only model cannot access the additional ordering and grouping cues existing in vector sketches and thus tends to struggle with sketches that have similar shapes but different category labels.
\Figure~\ref{fig:attention-visualize} visualizes the attention maps by our method and the ones encoding sketching order (used in Attention-using-Sketching-Order in Table~\ref{tab:ablationTUBerlin}).
It is observed that our attention maps estimated by RNN share a certain degree of similarity with the ones using sketching order, but the attention magnitudes by RNN are more adaptively biased. 

% ------ Figure ------> 
\begin{figure}[h]
    \centering
    \includegraphics[width=\linewidth]{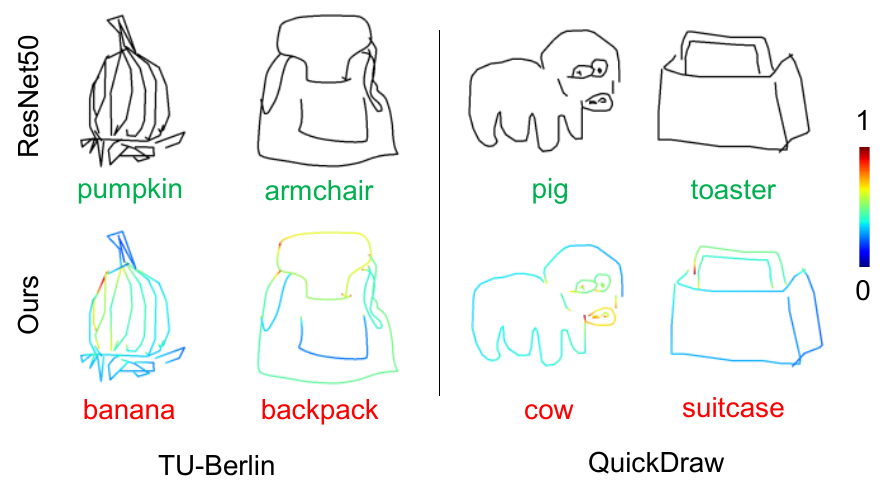}
    \caption{Recognition failures of our \OurNet{} (ResNet50).}
    \label{fig:cnnsucceed-ourfail}
\end{figure}

% <------ Figure ------ 

%=================================================================================
\textbf{Limitation.}
As shown in \Figure~\ref{fig:cnnsucceed-ourfail}, in some cases, the RNN in \OurNet{} may fail to produce correct attention guidance for the subsequent CNN, leading to recognition failures (e.g., the pumpkin), 
{possibly due to the inability in extracting effective sequential features from inputs with similar temporal ordering and grouping cues as other training sketches in different categories.}
Some sketches that are seemingly with ambiguous categories (e.g., the toaster) may also pose challenges to our method.
It is expected that human would make similar mistakes on such cases.
One possible solution to address the ambiguity is to put the sketched objects in context (i.e., scenes), and integrate our method with the context-based recognition methods~\cite{Zhang:2018:CSC:3229147.3229154,Zou_2018_ECCV}.

\section{Conclusion}
\label{sec:Conclusion}

In this work, we have proposed a novel single-branch attentive network architecture named \OurNet{} for vector sketch recognition. 
Our RNN-Rasterization-CNN design consistently improves the recognition accuracy of CNN-only models by 1-2\% on two existing large-scale sketch recognition benchmarks.
The key enabler for joining RNN and CNN together is a novel differentiable neural line rasterization module that performs in-network vector-to-raster sketch conversion.
Applying \OurNet{} to other tasks like sketch retrieval or sketch synthesis that need descriptive line-drawing features could be interesting to explore in the future.

{\small
\bibliographystyle{ieee}
\bibliography{reference}
}

\end{document}